\documentclass[letterpaper]{article} 
\usepackage{aaai2026}
\usepackage{times}  
\usepackage{helvet}  
\usepackage{courier}  
\usepackage[hyphens]{url}  
\usepackage{graphicx} 
\usepackage{natbib}  
\usepackage{caption} 
\usepackage{algorithm}
\usepackage{algorithmic}

\usepackage{newfloat}
\usepackage{listings}
\DeclareCaptionStyle{ruled}{labelfont=normalfont,labelsep=colon,strut=off} 
\floatstyle{ruled}
\newfloat{listing}{tb}{lst}{}
\floatname{listing}{Listing}
\usepackage{amsfonts}
\usepackage{amssymb}
\usepackage{amsmath}
\usepackage{amsthm}
\usepackage{xcolor}
\usepackage{stmaryrd}
\usepackage{mathtools}

\DeclareMathOperator*{\argmax}{arg\,max}

\definecolor{firebrick}{RGB}{178,34,34}

\title{Representation Understanding via Activation Maximization}
\author {
    Hongbo Zhu, Angelo Cangelosi
}
\affiliations {
    Manchester Centre for Robotics and AI, The University of Manchester\\
    \texttt{\{hongbo.zhu, angelo.cangelosi\}@manchester.ac.uk}
}

\usepackage{bibentry}

\nocopyright
\begin{document}

\maketitle

\begin{abstract}
Understanding internal feature representations of deep neural networks (DNNs) is a fundamental step toward model interpretability. Inspired by neuroscience methods that probe biological neurons using visual stimuli, recent deep learning studies have employed Activation Maximization (AM) to synthesize inputs that elicit strong responses from artificial neurons. In this work, we propose a unified feature visualization framework applicable to both Convolutional Neural Networks (CNNs) and Vision Transformers (ViTs). Unlike prior efforts that predominantly focus on the last output-layer neurons in CNNs, we extend feature visualization to intermediate layers as well, offering deeper insights into the hierarchical structure of learned feature representations. Furthermore, we investigate how activation maximization can be leveraged to generate adversarial examples, revealing potential vulnerabilities and decision boundaries of DNNs. Our experiments demonstrate the effectiveness of our approach in both traditional CNNs and modern ViT, highlighting its generalizability and interpretive value.
\end{abstract}

\section{Introduction}

Interpreting the internal representations of DNNs is fundamental to improving their transparency and trustworthiness. Among various techniques, AM offers a direct means to probe what a neuron or channel has learned by synthesizing inputs that maximize its activation. These optimized inputs, often referred to as Activation Maximization Signals (AMS), aim to reveal the underlying features to which neurons are most responsive.  A simple strategy for estimating neuron preferences involves identifying real input samples from the dataset that produce high activations. However, this approach is limited by data coverage and scalability. Neurons often respond to diverse or abstract concepts not fully represented in natural images \cite{nguyen2016multifaceted}. Furthermore, aggregating multiple high-activation samples does not guarantee a coherent or interpretable summary of the neuron's function.  

To address these limitations, feature visualization \cite{olah2017feature} techniques optimize the input directly without relying on real data to generate synthetic stimuli that activate targeted neurons. These methods often incorporate regularization, such as frequency-domain constraints or image augmentations to enhance interpretability and mitigate adversarial artifacts. Despite progress, most prior work focuses on the model output layer neurons, overlooking rich representations in intermediate layers. Moreover, generalizing AM across different architectures, such as ViT \cite{dosovitskiy2020image}, remains underexplored. Here, we propose a unified framework for representation understanding via activation maximization. Extending feature visualization to intermediate layers and covering both CNNs and ViT. We further explore how AM can generate adversarial examples, offering deeper insight into model behavior across architectures.

\begin{figure}
    \centering
    \includegraphics[width= 0.8\columnwidth]{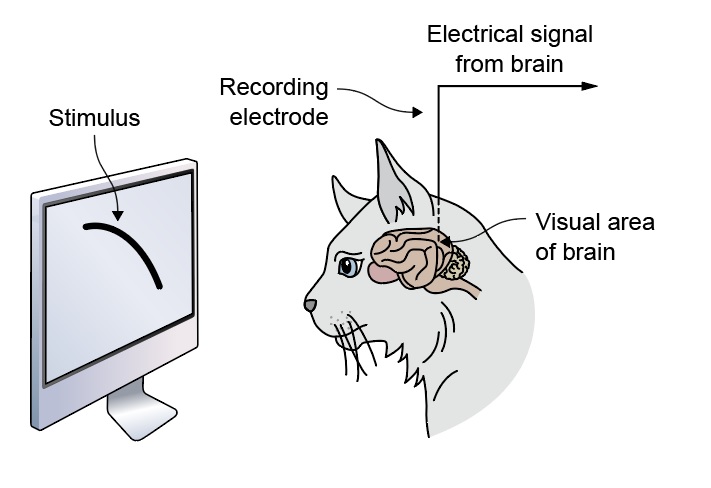}
    \caption{In the classic neuroscience experiment, \citet{hubel1959receptive} discovered a cat’s visual cortex neuron that fires strongly and selectively for a light bar when it is in certain positions and orientations.}
    \label{fig:neuroscience}
\end{figure}

\section{Related Works}
Understanding the internal representations of DNNs remains a core objective in the field of explainable AI. Among the various interpretability techniques, Activation Maximization holds great promise for uncovering the features learned by neurons or channels. This section reviews relevant literature on AM and its connection to adversarial robustness.

Activation Maximization was first proposed by \citet{erhan2009visualizing} as a technique for synthesizing input patterns that maximally activate specific neurons. This approach provided early insights into the internal representations of shallow networks. However, the initial visualizations often suffered from unnatural artifacts, prompting the development of improved optimization strategies and regularization techniques. For example, \citet{olah2017feature} introduced frequency-based priors, data augmentation methods, and total variation regularization to produce more interpretable and natural-looking visualizations and evaluated their method on GoogLeNet \cite{szegedy2015going}. These improvements enabled clearer insights into the semantic information captured by individual units across different layers. To address the multi-concept nature of neurons, \citet{nguyen2016multifaceted} introduced Multifaceted Feature Visualization (MFV), which generates distinct visualizations corresponding to the diverse feature types that activate a given neuron. This approach demonstrated that high-level units often encode multiple concepts, such as different instances or views of an object class, thereby offering a more comprehensive understanding of the neuron's function. More recently, \citet{fel2023unlocking} proposed Magnitude Constrained Optimization (MACO), which separates the optimization of phase and magnitude components in the Fourier domain. By constraining the magnitude to that of natural images and optimizing only the phase. 

Visualization techniques have been extensively developed for CNNs. \citet{zeiler2014visualizing} introduced a deconvolutional approach for projecting intermediate feature activations back into pixel space, enabling layer-wise interpretation. Complementary approaches, such as network inversion \cite{mahendran2015understanding} and network dissection \cite{bau2017network}, aimed to reconstruct input images from activations or to map neurons to human-interpretable semantic concepts. While CNN-based visualization has matured, applying similar interpretability methods to ViTs remains challenging due to their token-based attention mechanisms and lack of spatial inductive bias. Most early AM studies focused primarily on output-layer neurons. However, intermediate layers encode hierarchical abstractions that are critical for understanding the model’s internal processing.

In addition to interpretability, AM techniques also intersect with research on adversarial robustness. \citet{szegedy2013intriguing} demonstrated that imperceptible perturbations to input images could cause DNNs to misclassify them—a phenomenon now known as adversarial examples \cite{goodfellow2014explaining}. Notably, such perturbations can be generated via gradient ascent similar to those used in AM, suggesting a deep connection between representation understanding and adversarial attack methodologies. 

\begin{figure*}
    \centering
    \includegraphics[width= 0.90\textwidth]{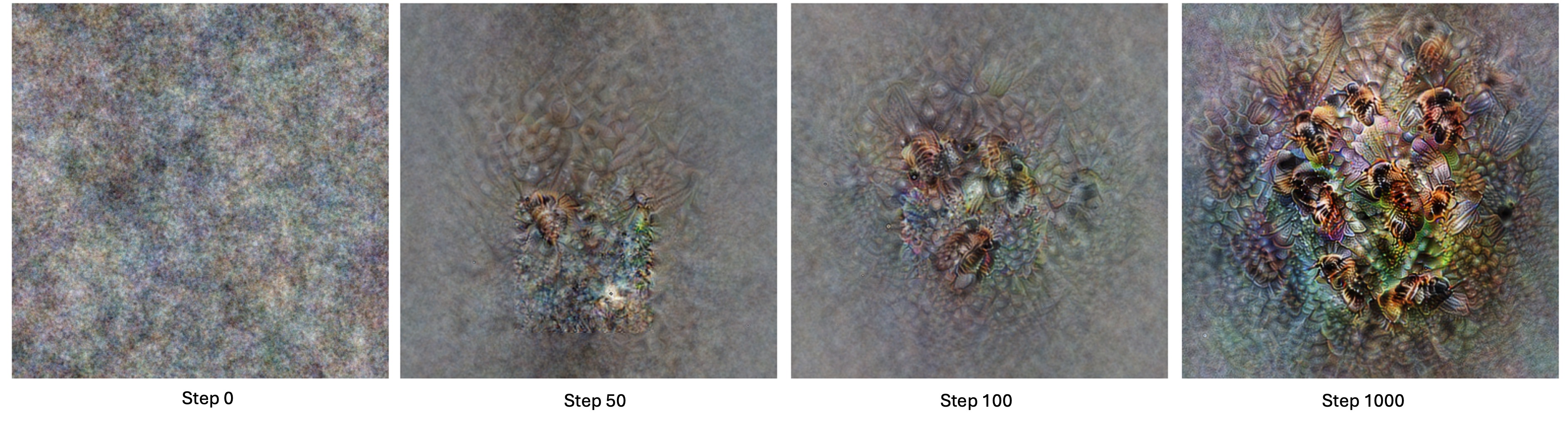}
    \caption{Starting from a random initialization in the Fourier domain, we iteratively optimize it with gradient ascent to maximize the activation of an objective logits neuron (Bee) on MobileNet.}
    \label{fig:optimization_process}
\end{figure*}

\section{Methodology}
\subsection{Feature Visualization Via Activation Maximization}
We restricted ourselves needlessly to searching for an input pattern from the training or test sets; instead, we can take a more general view of treating maximizing the activation of a unit (neuron or channel) as an optimization problem. Let $f$ denote our neural network and let $f^{l}_{t}(x)$ be the activation of a given unit $t$ from layer $l$ in the network of an input sample $x$. Assuming a fixed $f$ (for instance, the parameters after training the network), we can view our idea as looking for 
\begin{equation}
    x^* = \argmax_x(f^{l}_{t}(x))
\label{eq:1}
\end{equation}

This is, in general, a non-convex optimization problem. However, we can attempt to find a local maximum using gradient ascent. This can be achieved by performing gradient ascent in the input space, i.e. computing the gradient of $f^{l}_{t}(x)$ and moving $x$ in the direction of this gradient, so as to maximize $f^{j}_{t}(x)$.

\begin{equation}
    x \leftarrow x+\eta \frac{\partial f^{l}_{t}(x)}{\partial x}
\label{eq:2}
\end{equation}
This optimization technique (activation maximization) is applicable to any network in which we can compute the above gradients. Like any gradient descent \cite{ruder2016overview} technique, it does involve a choice of hyperparameters: the learning rate and a stopping criterion (the maximum number of gradient ascent updates) \cite{erhan2009visualizing}. Note that this gradient ascent process is similar to that of the gradient descent used to train neural networks via backpropagation, except that here we optimize the network input $x$ instead of the network parameters $\theta$, which are frozen. Optimization can give us an example input that causes the desired behaviour; it separates the things causing behavior from things that merely correlate with the causes.

\subsection{Optimization Challenge In Pixel Domain}

In the pixel domain, we attempt to optimize an input image \( x \in \mathbb{R}^{H \times W \times C} \) by maximizing the activation of a target neuron using the objective in Eq.\ref{eq:1}, and updating \( x \) through gradient ascent as described in Eq.\ref{eq:2}. However, direct optimization in pixel space often leads to the emergence of noisy, high-frequency artifacts in the synthesized images. This is largely due to the structure of the gradient \( \frac{\partial f}{\partial x} \), which frequently exhibits large components in high-frequency directions. 
Neural networks are known to be sensitive to such small, high-frequency perturbations \cite{goodfellow2014generative}, a property that underlies the phenomenon of adversarial examples.
Unlike natural images, which predominantly contain low-frequency content, the optimized inputs often exploit peculiar correlations between input features and network activations that may not be perceptible to humans. Prior work has shown that deep networks can generalize well by utilizing input information—including local textures, color statistics, edge orientations, and unintuitive high-frequency features—that might be unrecognizable to human observers~\citep{yin2019fourier}. As a result, optimizing in the raw pixel space without any regularization or constraints typically yields unnatural images that strongly activate the target neurons but lack human interpretability. These images resemble ``neural illusions''—inputs that do not occur in the natural data distribution, but which the network nonetheless responds to with high confidence. The emergence of these adversarial-like patterns indicates that, without appropriate priors or constraints, activation maximization in pixel space tends to uncover directions in the input space that lie far outside the manifold of natural images.
This insight highlights the need for incorporating priors, regularizers, or alternative parameterizations to guide the optimization process toward more semantically meaningful and interpretable visualizations, rather than mere artifacts that exploit the network’s vulnerabilities.


\begin{algorithm}
    \caption{Feature-Vis}
    \label{alg:fv}
\begin{algorithmic}[1] 
    \STATE {\bfseries Require} $ z_0 = a_0+b_0i \sim \mathbb{C}^{W \times H}$
    \FOR{$i \in \llbracket 0, N-1 \rrbracket$}
        \STATE  \( \tau \sim \mathcal{T} \) (e.g., jittering, scaling, rotation)
        \STATE $x_i = (\tau \circ \mathcal{F}^{-1})(z_i)$ 
        \STATE $z_{i+1} = z_i + \eta \nabla_{z_i} f^{l}_{t}(x_i)$
    \ENDFOR
    \STATE {\bfseries Return} $x^* = \mathcal{F}^{-1}(z_N)$
\end{algorithmic}
\end{algorithm}

\subsection{Optimization In Frequency Domain}

\begin{figure}[h]
    \centering
    \includegraphics[width=0.9\columnwidth]{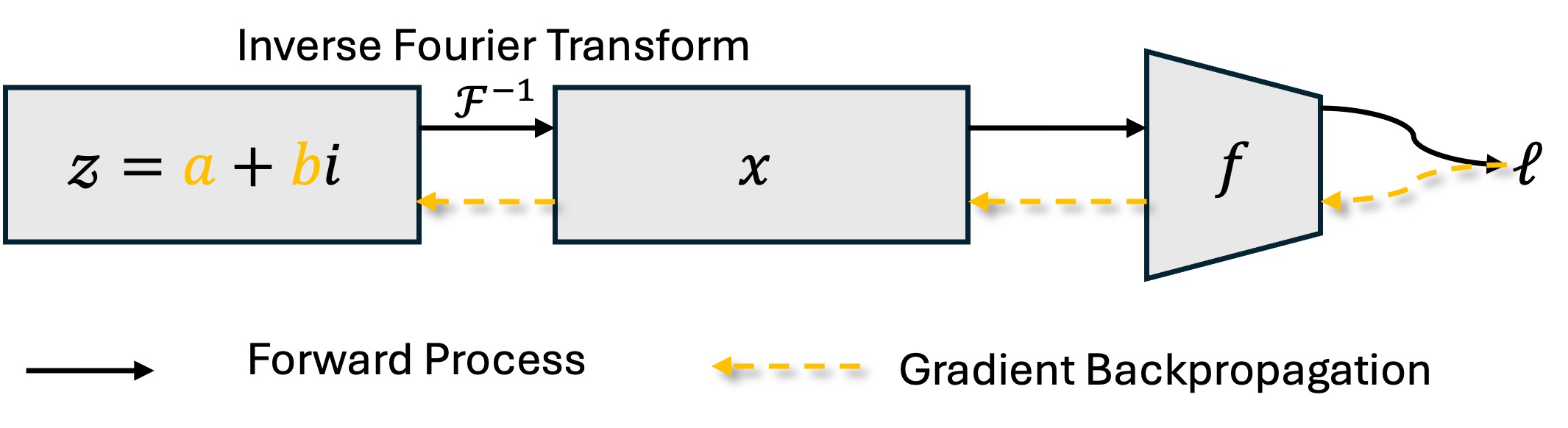}
    \caption{Visualization pipeline in Fourier domain.}
    \label{fig:fv_pipline}
\end{figure}

One of the central challenges in feature visualization is mitigating the presence of high-frequency noise, which often leads to unnatural or adversarial visual artifacts. To generate interpretable and human-perceivable visualizations, it is essential to introduce structural priors or constraints that guide the optimization process toward more natural image distributions—such as photo-realistic images or those resembling training data statistics~\cite{nguyen2019understanding}.
A promising solution involves parameterizing the optimization process directly in the frequency domain \cite{cai2021frequency}. Rather than optimizing over pixel values in the spatial domain, we represent the image \( x \) as the inverse Fourier transform \cite{heckbert1995fourier} of a complex-valued tensor:
\[x = \mathcal{F}^{-1}(z), \quad z = a + b i \in \mathbb{C}^{W \times H}\]
where \( a \) and \( b \) denote the real and imaginary components of the frequency representation respectively. This formulation inherently suppresses high-frequency artifacts by encouraging smooth and coherent spectral structures. 
As illustrated in Figure~\ref{fig:fv_pipline}, the forward process begins with transforming \( z \) into the image space via Inverse Fourier Transform \( \mathcal{F}^{-1} \), followed by a loss computation. The gradients are then backpropagated through this entire pipeline to update \( z \) accordingly. This approach allows the optimization to operate over the entire frequency spectrum while regularizing against adversarial artifacts.
Algorithm~\ref{alg:fv} summarizes this procedure. The process begins with an initial random complex tensor \( z_0 \sim \mathbb{C}^{W \times H} \). For each iteration \( i \in \{0, \dots, N{-}1\} \) \footnote{We fix the total optimization step of N=1000 across all the feature visualization experiments in this paper.}, a transformation \( \tau \) is sampled and applied to the image \( x_i \), computed as: \(x_i = (\tau \circ \mathcal{F}^{-1})(z_i)\). A objective function \( f_t^l \), targeting specific activations $t$ in layer $l$ of the network, is evaluated on \( x_i \). The gradient of this loss with respect to \( z_i \) is used to perform an update step:
\[z_{i+1} = z_i + \eta \nabla_{z_i} f_t^l(x_i)\]
where \( \eta \) is the learning rate. After \( N \) iterations, the final image is reconstructed by computing: \(x^* = \mathcal{F}^{-1}(z_N)\), this optimization process is illustrated in Figure \ref{fig:optimization_process}.
This frequency-based parameterization not only suppresses adversarial high-frequency noise but also modifies the optimization landscape, potentially altering the basins of attraction. The result visualizations exhibit enhanced naturalness and semantic coherence, contributing to more interpretable feature attribution.

\subsection{Optimization Objectives} 
\begin{figure}[h]
    \centering
    \includegraphics[width=0.99\columnwidth]{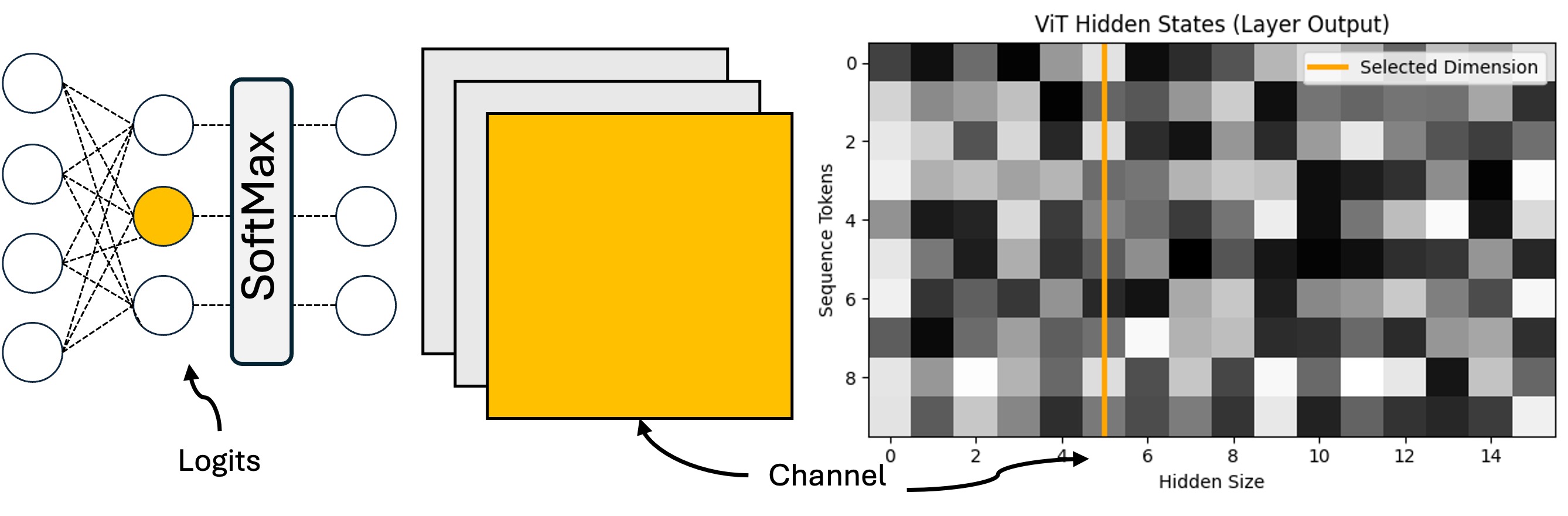}
    \caption{Visualization objectives schematic diagram.}
    \label{fig:objectives}
\end{figure}

In this work, our visualization objectives are unified across both CNNs and ViT, targeting (1) the penultimate layer logits (Figure \ref{fig:objectives} leftmost) and (2) the intermediate layer channels (Figure \ref{fig:objectives} middle and right). For CNNs, we visualize the penultimate logits layer neurons to capture class-specific activation patterns and also focus on individual channels in intermediate convolutional layers, which often correspond to localized semantic features. In ViT, we similarly target the penultimate logits layer, as well as specific hidden dimensions in intermediate transformer layers, which we treat as feature channels. For a given hidden dimension index in ViT, we extract activations using \texttt{hidden\_states[layer][\,:\, ,\,:\, ,\,idx]} and aggregate them across all tokens to guide input optimization. This channel-wise visualization approach, inspired by the correspondence between CNN channels and transformer hidden dimensions, allows us to identify input patterns that elicit strong activations and to analyze the hierarchical and semantic representations captured by both architectures. 

\begin{figure*}
    \centering
    \includegraphics[width=0.9\textwidth]{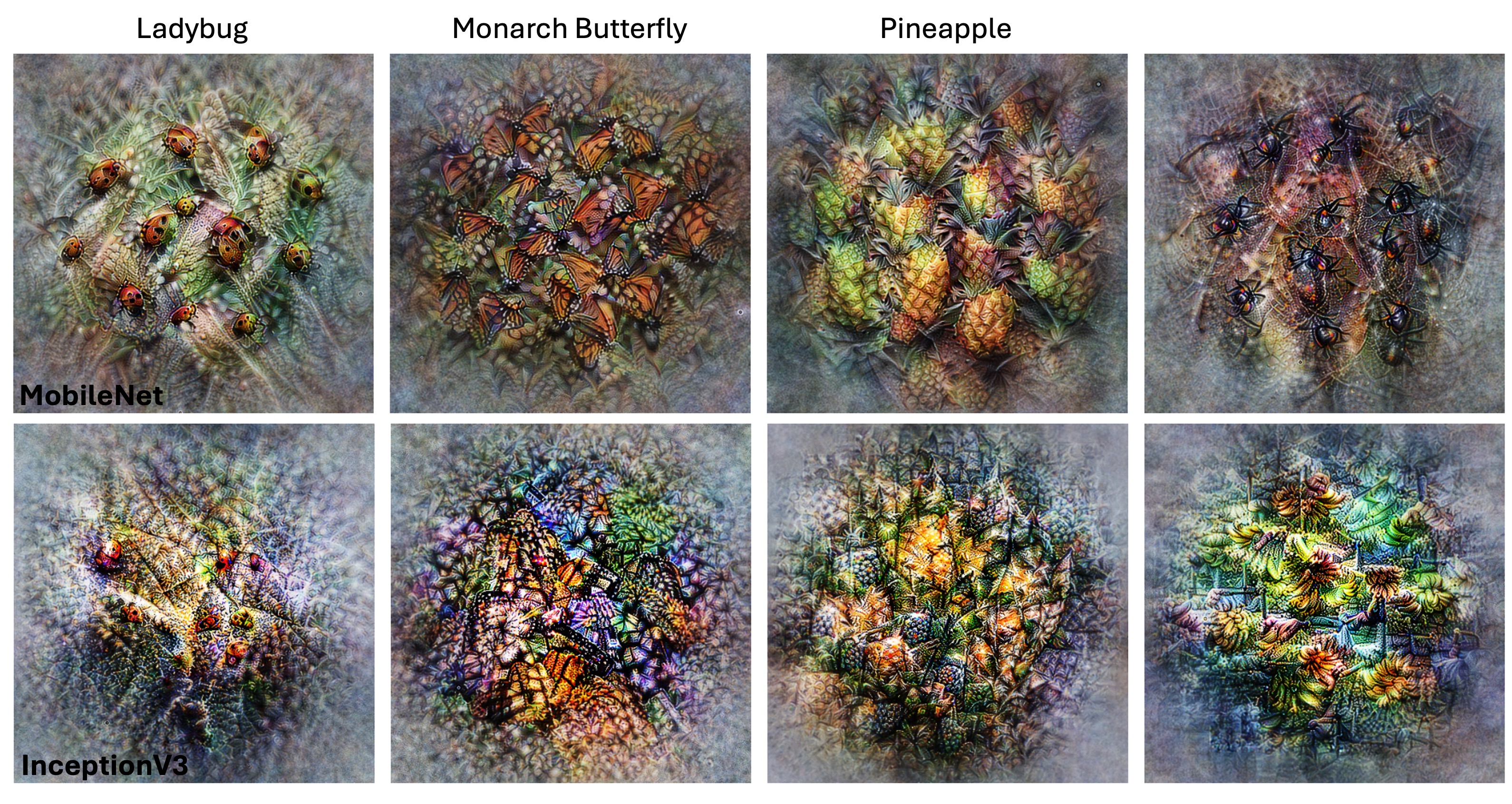}
    \caption{Visualization results of logits neurons on MobileNet and InceptionV3, the first three columns visualizations corresponding to Ladybug, Monarch Butterfly, and Pineapple class neurons, while the last column neurons are randomly selected from other classes.}
    \label{fig:logits_up}
\end{figure*}

\section{Experiments}

\paragraph{Experiment Setting}
We perform a comprehensive experiment across three different scale CNN Models: MobileNet \cite{howard2017mobilenets}, InceptionV3 \cite{szegedy2016rethinking}, ResNet50V2 \cite{he2016deep}, and one Transformer-based ViT-B/16, starting from visualizing the last logits layer neurons to the intermediate layer channels. Finally, we propose an easy-to-implement but effective AM-based adversary example generation method and evaluate it on both ResNet50V2 and ViT-B/16.

\subsection{On Last Logits Layer}

We chose the penultimate layer before the softmax, not the layer after the softmax, as maximizing the target class probability doesn't necessarily guarantee that the activation value of the corresponding neuron will also be maximized.

\[\text{Softmax}(z_i) = \frac{e^{z_i}}{\sum_{j=1}^{n} e^{z_j}} = \frac{e^{z_i}}{e^{z_i}+\sum_{j\neq i} e^{z_j}}
\]

$\text{Softmax}(z_i)$ yields a high probability does not necessarily imply that the corresponding logit $z_i$ is large; it may also be because $\sum_{j\neq i} e^{z_j}$ is relatively small.

\begin{table}
\small
\centering
\begin{tabular}{lcccc}
\hline
 & \textbf{MobileNet} & \textbf{InceptionV3} & \textbf{ResNet50V2} & \textbf{ViT-B/16} \\
\hline
\textbf{Par} & $\sim$4.2M & $\sim$23.9M & $\sim$25.6M & $\sim$86M \\
\hline
\end{tabular}
\caption{Comparison of model sizes in terms of parameters.}
\label{tab:model_params}
\end{table}

To gain deeper insights into how different neural network architectures represent semantic categories, we applied our proposed method to visualize the individual pre-softmax layer neurons. Figure~\ref{fig:logits_up} and 8 (Appendix) present the optimized visualizations for three semantically distinct class neurons: Ladybug, Monarch Butterfly, and Pineapple, as well as a control set of neurons randomly selected from other categories. For all models, the first three columns of the visualization exhibit high-level semantic features that align with their respective classes. For instance, all four models generate salient red-and-black spotted patterns for the Ladybug neuron, textured orange-and-black wings for the Monarch Butterfly, and clustered yellow-green motifs for Pineapple. This suggests that neurons at the pre-softmax layer capture class-specific global structures rather than merely local texture patterns.

The visualizations reveal key differences in how various architectures compose their internal representations. MobileNet tends to generate repeated motif clusters—such as multiple instances of ladybugs or butterflies—indicating a strong reliance on local features. InceptionV3 produces more spatially coherent and colorful patterns, likely benefiting from its multi-scale receptive fields and heterogeneous architecture. ResNet50V2 emphasizes contrastive and sharply segmented forms, suggesting robust edge-aware activations facilitated by residual connections. In contrast, ViT-B/16 generates more abstract and diffuse visualizations, lacking clear spatial boundaries, which reflects its reliance on global self-attention rather than localized filters. Overall, CNNs exhibit more interpretable and localized patterns aligned with prototypical object structures, a result of their spatial inductive biases such as translation invariance and hierarchical feature composition \cite{luo2016understanding, kayhan2020translation}. ViTs, on the other hand, capture high-level semantics in a more distributed and holistic manner \cite{vaswani2017attention}, leading to globally integrated but less visually localized representations.

\begin{figure*}[h]
    \centering
    \includegraphics[width= 0.95\textwidth]{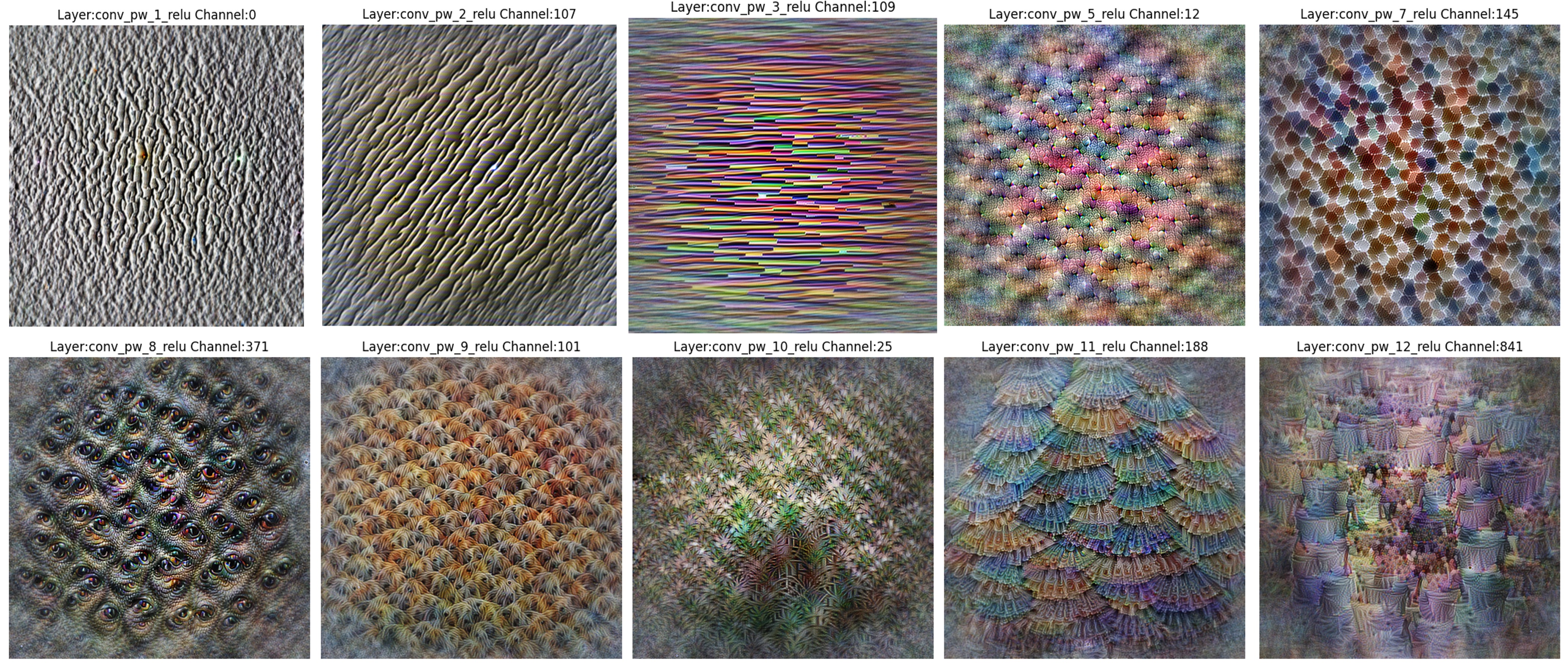}
    \caption{Feature visualization on MobileNet across different intermediate layers channels.}
    \label{fig:mobilenet}
\end{figure*}

\subsection{On Intermediate Layers}
The visualization results of intermediate channels reveal distinct characteristics of the learned representations across layers and model architectures.

\paragraph{MobileNet}
Figures \ref{fig:mobilenet} and 10 (Appendix) illustrate the feature visualizations for the intermediate representation of MobileNet. This reveals the nature of learned representations at different depths of the network and shows a clear progression from low-level edge and texture detectors to high-level abstract and semantic patterns and objects, which is consistent with hierarchical feature learning in CNNs. 
Early layers (conv\_pw 1 to 3) predominantly highlight edge-like structures, oriented gradients, and fine textures, with many channels exhibiting Gabor-like waveforms \cite{mehrotra1992gabor}, suggesting orientation- and frequency-specific tuning akin to classical edge detectors. These encoded spatial orientation, frequency, and direction features are necessary for subsequent abstraction. For middle layers (conv\_pw 5 to 8), the activations evolve to capture more complex and periodic patterns, including geometric tiling (e.g., honeycombs, grids, hexagonal motifs), checkerboards, and spirals. Such responses indicate tuning to combinations of edges and textures, likely corresponding to object parts or biologically relevant textures.  At higher layers (conv\_pw 9 to 12), the activations reflect increasingly semantic and structured representations, including object-part-like textures (e.g., fur, eyes), symmetrical and radial motifs (suggestive of flowers or faces), and complex interleaved forms resembling animal figures, plant-like structures. These layers appear to encode high-level semantic prototypes relevant to object categorization. Across all layers, a clear progression in representational complexity is evident—from simple, low-level features to structured, domain-relevant abstractions. Additionally, the recurrent appearance of repetitive tiling in mid and deep layers may contribute to pose and scale invariance, while the presence of biologically inspired patterns aligns with the hypothesis that CNNs, when trained on natural imagery, learn features that mirror structures found in the natural world.

\paragraph{ResNet50V2}
Figures 11 and 12 (Appendix) illustrate the visualization results for ResNet50V2, revealing its distinctive representational characteristics in comparison to lightweight architectures such as MobileNet. Notably, ResNet50V2 demonstrates greater global coherence and structural abstraction, particularly from its deeper residual blocks, where activations emphasize coherent shape primitives and semantically meaningful object parts. This behavior reflects the compositional nature of residual learning, wherein skip connections \cite{zhou2019unet++} not only preserve low-level textures but also facilitate the integration of increasingly abstract features. Such progressive refinement enables the model to construct rich hierarchical representations, underpinning its strong performance across diverse vision tasks. As a deep convolutional neural network equipped with residual connections, ResNet50V2 benefits from improved gradient flow, allowing for the learning of more useful feature encodings. Compared to MobileNet, ResNet50V2 exhibits a deeper and more diverse representational hierarchy, with stronger semantic alignment and clearer modular organization of part-object structures. These observations support the hypothesis that residual architectures not only retain fine-grained features but also promote the emergence of interpretable and discriminative visual abstractions, highlighting the scalability and expressiveness of deep residual models.

\begin{figure*}
    \centering
    \includegraphics[width= 0.95\textwidth]{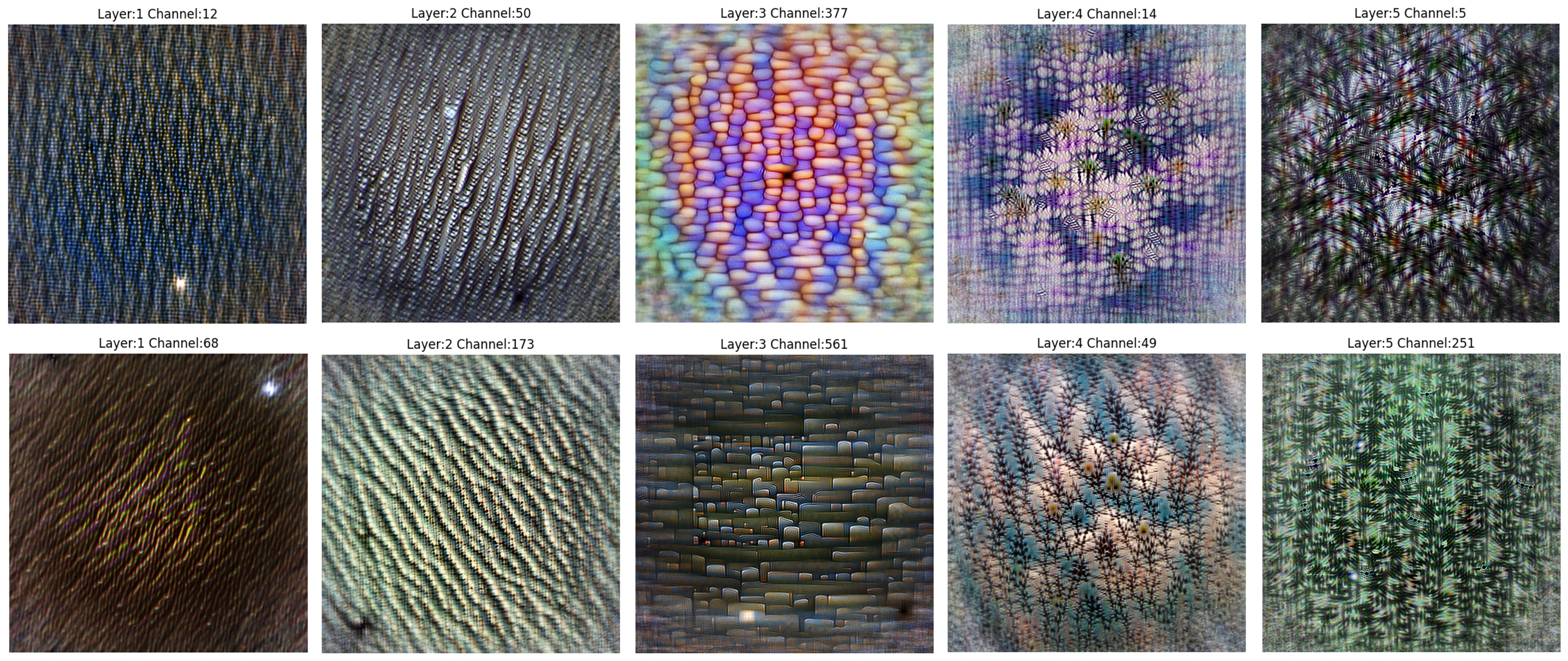}
    \caption{Feature visualization for ViT model across different intermediate layers.}
    \label{fig:vit}
\end{figure*}

\paragraph{ViT-B/16}
Figure \ref{fig:vit} illustrates feature visualizations from various shallow layers (1–5) of a standard ViT, revealing a progressive evolution in representational abstraction distinct from that observed in CNNs. Unlike CNNs, which rely on localized spatial filters and exhibit strong inductive biases \cite{goyal2022inductive} such as translation equivariance, ViTs leverage attention-based, non-local token interactions that result in more dispersed and global activation patterns. In the earliest layers (1–2), ViT tends to capture fine-grained, high-frequency textures, such as periodic dots, lines, or directional stripe patterns, reminiscent of pixel arrays or fabric-like structures. These activations likely encode low-level texture details, token-grid alignment, and spatial consistency, despite the absence of convolutional operations. In layers 3–4, the representations shift toward more structured, compositional groupings that suggest emerging inter-token relationships and positional encoding mechanisms. By Layer 5, activations become markedly complex and globally organized, reflecting the integration of long-range dependencies, hierarchical structures, and latent object-level abstractions made possible by multi-head self-attention and deep token mixing. These observations support the hypothesis that ViTs, through their attention mechanisms, are capable of capturing holistic and semantically rich representations earlier in the network compared to their convolutional counterparts.

\subsection{Representation Comparison Between CNN and ViT}
CNNs, such as MobileNet and ResNet50V2, typically construct representations in a bottom-up manner, starting with edges and textures in early layers and gradually composing object parts and semantics in deeper layers through progressively growing receptive fields. MobileNet emphasizes local, generating sharper and texture-focused features, whereas ResNet50V2 strikes a balance between local detail and global abstraction. In contrast, ViT diverges from this paradigm by utilizing self-attention mechanisms that enable non-local interactions from the outset, resulting in globally integrated features even in early or shallow layers. However, this early-stage abstraction often leads to representations that are less readily interpretable to humans. Despite these differences, both CNNs and ViT demonstrate a form of progressive specialization across layers, CNNs through localized filters and ViT through attention-based token mixing. These findings underscore the complementary nature of architectural biases and their impact on representation learning. 
It seems there is a trend: larger models, such as ViT-B/16 (86M), tend to produce more abstract feature representations compared to smaller models like MobileNet (4.2M). The model size can be seen from Table \ref{tab:model_params}. This increase in abstraction often complicates interpretability, making it more challenging to generate human-understandable visualizations, particularly in deeper layers.

\subsection{Activation Maximization for Adversarial Examples Generation}

Given a pre-trained model $f:\mathbb{R}^{H \times W \times 3} \rightarrow \mathbb{R}^{C}$  and an input image $x_\text{orig}$, we seek an adversarial example  $x_{\text{adv}}$ that maximizes the model’s logits output for a target class $t$,  while constraining the perturbation to be imperceptible and smooth. 
Specifically, we try to solve:

\[
x_{\text{adv}} = \arg\max_{x \in \mathcal{C}} \left( f_t(x) - \lambda \cdot \text{TV}(x - x_\text{orig}) \right)
\]

where $f_t(x)$ denotes the logits corresponding to the target class $t$,  $\text{TV}(\cdot)$ is the total variation regularizer \cite{strong2003edge} that encourages spatial smoothness in the perturbation, $\lambda$ is a hyperparameter controlling the trade-off between attack strength and imperceptibility, and $\epsilon$ limits the maximal pixel-wise perturbation magnitude. The constraint set $\mathcal{C}$ is defined as
\[
 \mathcal{C} = \left\{x \in [0,1]^{H \times W \times 3} \;\middle|\; \|x - x_{\text{orig}}\|_\infty \leq \epsilon \right\}
\]

To solve the above constrained maximization, we employ iterative gradient ascent with projected updates:

\[
x \leftarrow \Pi_{\mathcal{C}} \left( x + \alpha \cdot \nabla_{x} \left( f_t(x) - \lambda \cdot \text{TV}(x - x_\text{orig}) \right) \right)
\]

where $\alpha$ is the step size, and $\Pi_{\mathcal{C}}$  denotes projection onto the $\ell_\infty$ ball of radius $\epsilon$ centered at $x_\text{orig}$,  then clamping to the valid input range $[0, 1]$. Namely:
\begin{enumerate}
    \item $L_\infty$-ball projection:
    \[
        x \leftarrow x_{\text{orig}} + \mathrm{clip}\big(x - x_{\text{orig}},\, -\epsilon,\, \epsilon\big)
    \]
    \item Valid range projection: 
    \[
        x \leftarrow \mathrm{clip}(x, 0, 1)
    \]
\end{enumerate}

This formulation ensures that the adversarial example (i) maximizes the logits of the target class, (ii) yields a spatially smooth and imperceptible perturbation via the total variation regularizer, (iii) remains close to the original input under the $\ell_\infty$ constraint, and (iv) satisfies the valid pixel range.


\begin{table}
\centering
\begin{tabular}{ccccc}
\hline
\textbf{Model}&$\epsilon$& $\alpha$ & $\lambda $ & Step \\
\hline
\textbf{Resnet50V2} & 0.01 & 0.01 & 1e-4 & 30 \\
\hline
\textbf{ViT} & 0.05 & 0.01& 1e-4 & 30 \\
\hline
\end{tabular}
\caption{Hyperparameter setting for adversary example generation with activation maximization.}
\label{tab:adversary_hyperparams}
\end{table}

Figure 9 (Appendix) visually illustrates the effectiveness of the proposed AM-based approach for crafting targeted adversarial examples across three distinct demo images: ladybug, monarch butterfly, and pineapple. For each row of the image, we present the original input (leftmost), the computed perturbation (amplified $\times$10 for visibility), and the generated adversarial examples classified as a specific incorrect adversary class. Two sets of perturbation–adversarial image pairs are included per input to showcase robustness across different perturbation patterns and model targets. The qualitative results demonstrate that our method successfully induces misclassification into the specified target adversary class (e.g., ladybug $\rightarrow$ monarch butterfly, monarch butterfly $\rightarrow$ pineapple pineapple $\rightarrow$ ladybug), while preserving perceptual similarity with the original image. The perturbations generated for the ViT model show a distinct grid/checkerboard-like pattern, visible in the ``Perturbation (x10)" columns of each row. This pattern is notably absent or less structured in the perturbations for ResNet50V2. This is possibly due to the ViT's patch-based input encoding and lack of convolutional inductive bias. The use of different hyperparameters per model, as listed in Table~\ref{tab:adversary_hyperparams}. Notably, ViT requires a larger perturbation bound ($\epsilon=0.05$) to achieve comparable fooling rates, reflecting its increased robustness or different sensitivity profile relative to convolutional networks. In summary, these results validate the utility of activation maximization with TV regularization as an effective approach for generating targeted, imperceptible, and structured adversarial examples. This technique not only highlights critical weaknesses in current visual recognition models but also offers insights into the spatial structure of adversarial vulnerabilities.

\section{Conclusion}

In this work, we proposed a unified framework for representation understanding via AM, applicable across both CNNs and ViT. By extending AM beyond the output layer to intermediate layers and optimizing in the frequency domain, we achieved more interpretable and semantically meaningful visualizations, while also revealing structural differences in how various architectures encode information. Furthermore, we showed that AM can be adapted to generate targeted adversarial examples, offering a new perspective on the intersection of interpretability and robustness.

Despite these advances, several important challenges remain. The optimization process becomes increasingly difficult as model size grows, especially for large-scale ViTs, where activation landscapes are more complex and gradients tend to be less informative. Future work should explore more effective optimization strategies, potentially involving learned priors, generative models, or diffusion-based approaches to improve convergence and stability in large models. Another important direction is enhancing the human interpretability of synthesized visualizations. While current methods can highlight discriminative features, bridging the semantic gap between optimized patterns and human-recognizable concepts remains an open problem. Integrating human-in-the-loop feedback, or grounding in natural image distributions, may help produce more understandable results. Finally, a critical and largely unexplored area is understanding how individual neurons or channels interact and combine to form more complex representations. While AM reveals preferred stimuli for isolated units, studying compositionality and inter-neuronal dependencies could offer deeper insight into the functional organization of learned representations.


\bibliography{aaai2026}
\begin{figure*}
    \centering
    \includegraphics[width=0.93\textwidth]{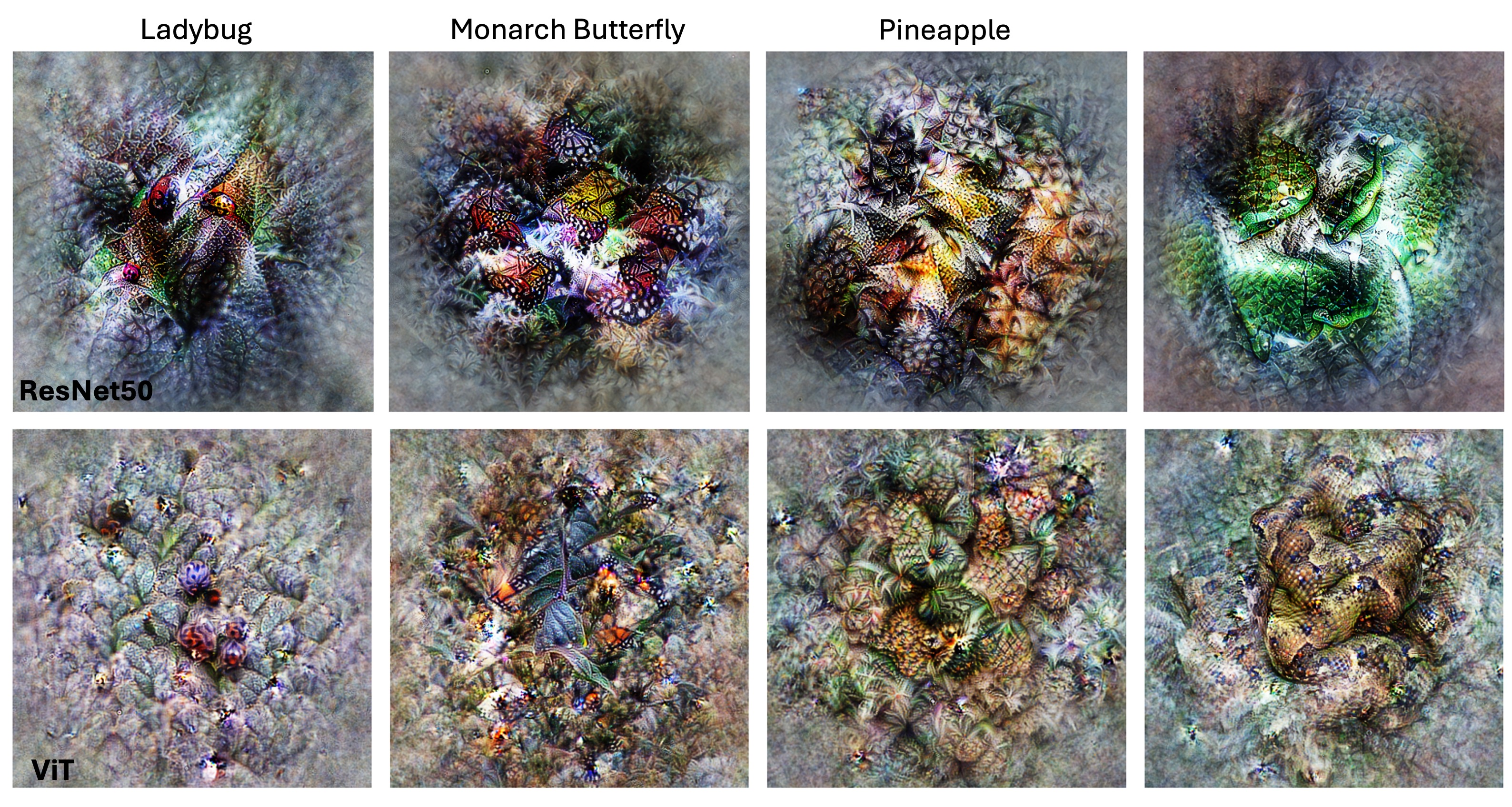}
    \caption{Visualization results of pre-softmax layer Logits neurons on ResNet50V2 and ViT-B/16.}
    \label{fig:logits_down}
\end{figure*}

\begin{figure*}
    \centering
    \includegraphics[width=0.90\textwidth]{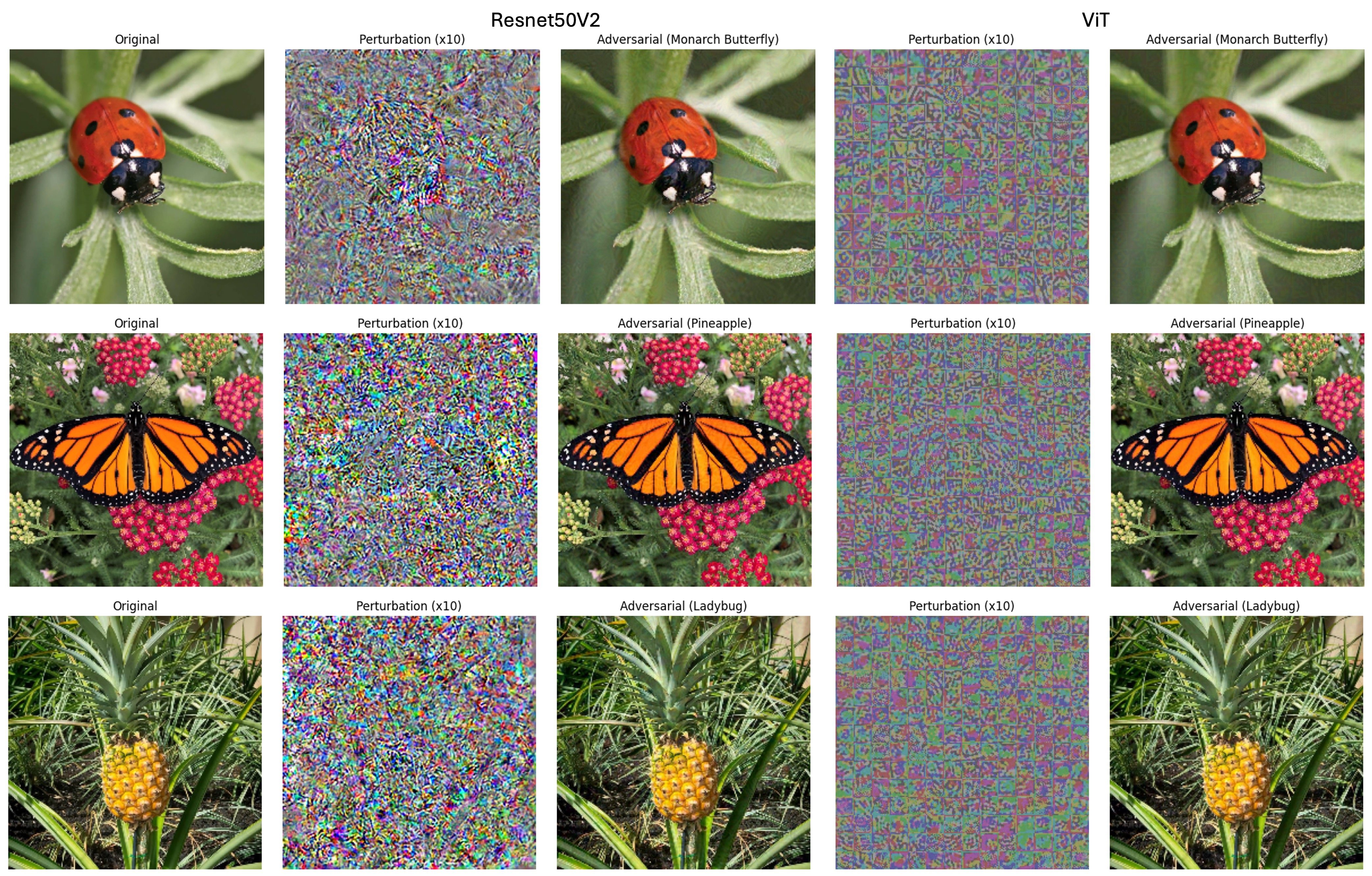}
    \caption{Adversary examples generated with our proposed class logits guided Activation Maximization optimization.}
    \label{fig:adversary}
\end{figure*}

\begin{figure*}
    \centering
    \includegraphics[width= 0.93\textwidth]{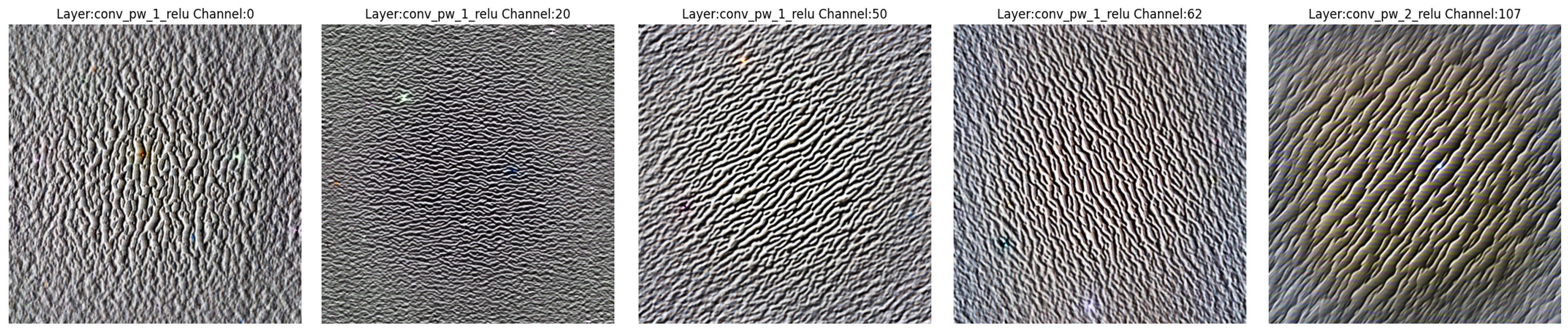}
    \includegraphics[width= 0.93\textwidth]{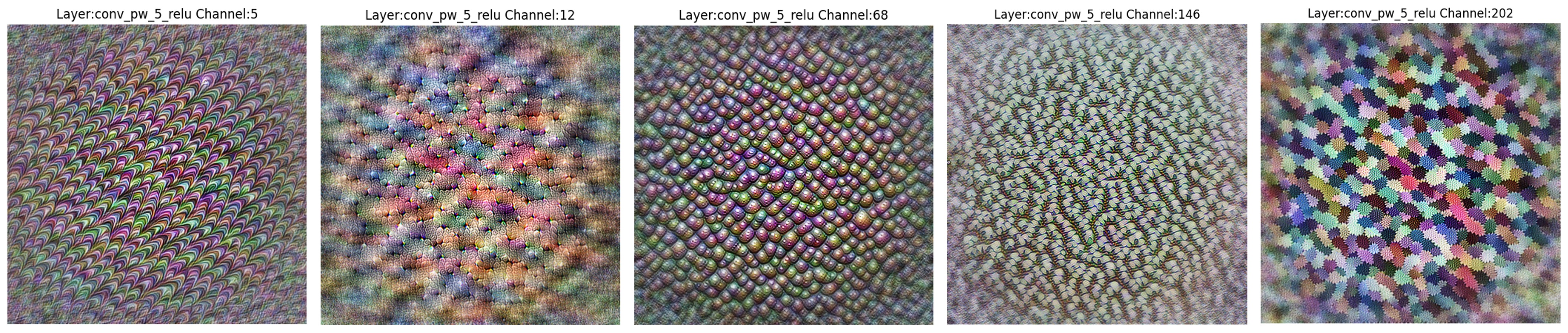}
    \includegraphics[width= 0.93\textwidth]{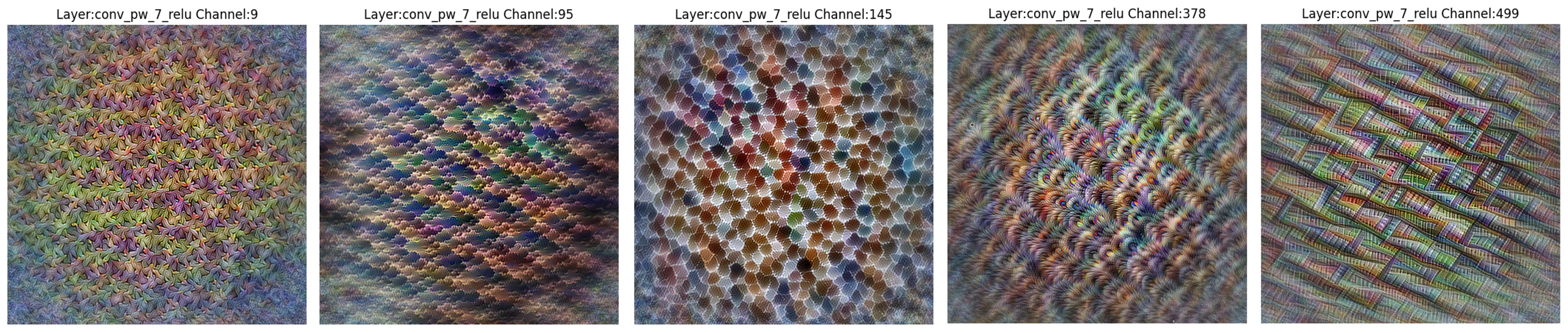}
    \includegraphics[width= 0.93\textwidth]{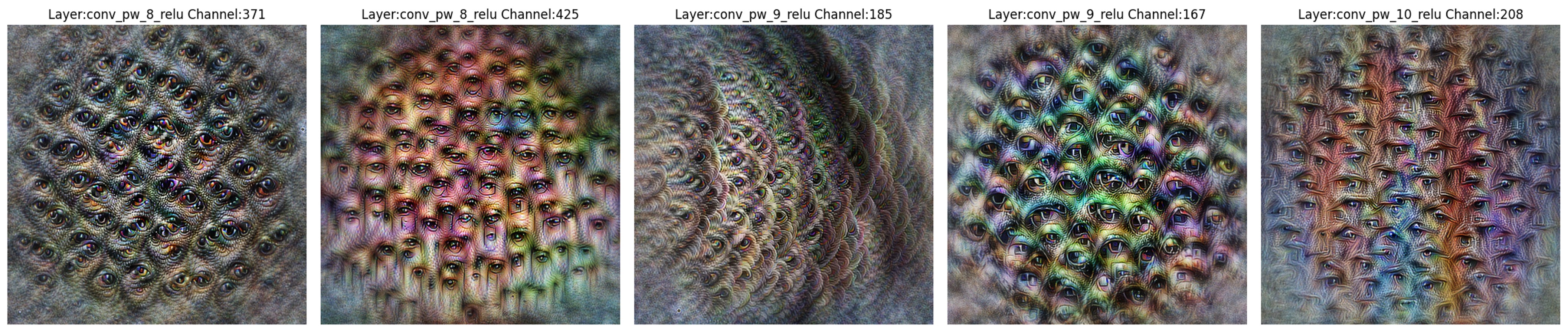}
    \includegraphics[width= 0.93\textwidth]{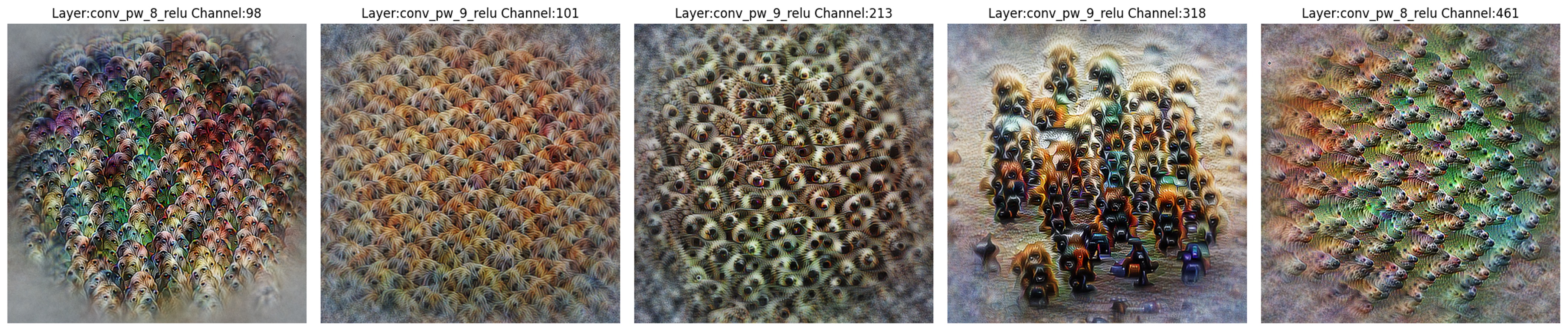}
    \includegraphics[width= 0.93\textwidth]{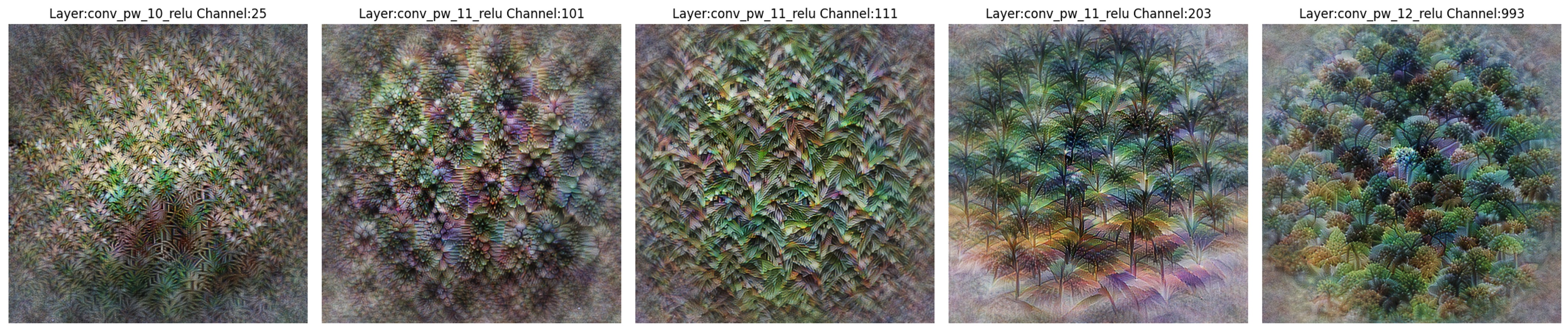}
    \includegraphics[width= 0.93\textwidth]{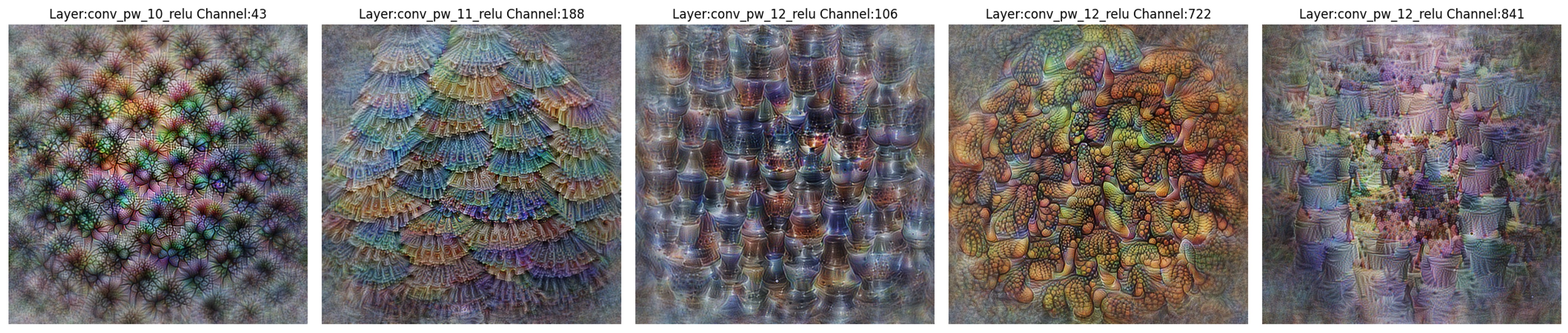}
    \caption{More visualization results of Mobilenet across different intermediate layers channels.}
    \label{fig:mobilenet_ap}
\end{figure*}

\begin{figure*}
    \centering
    \includegraphics[width= 0.93\textwidth]{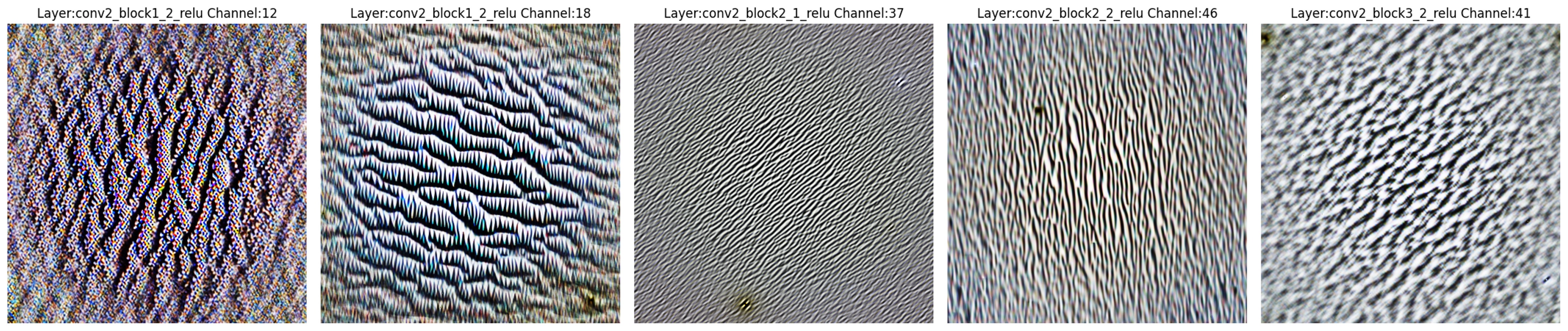}
    \includegraphics[width= 0.93\textwidth]{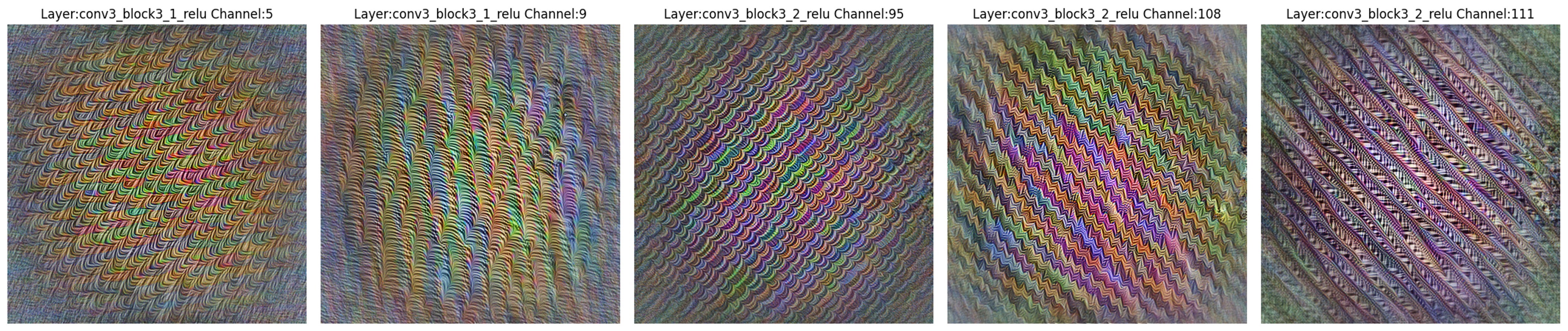}
    \includegraphics[width= 0.93\textwidth]{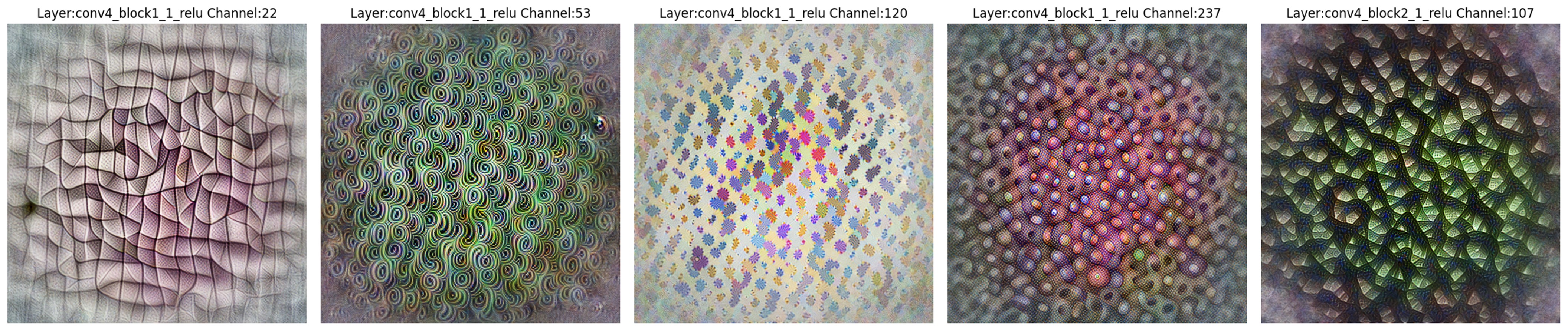}
    \includegraphics[width= 0.93\textwidth]{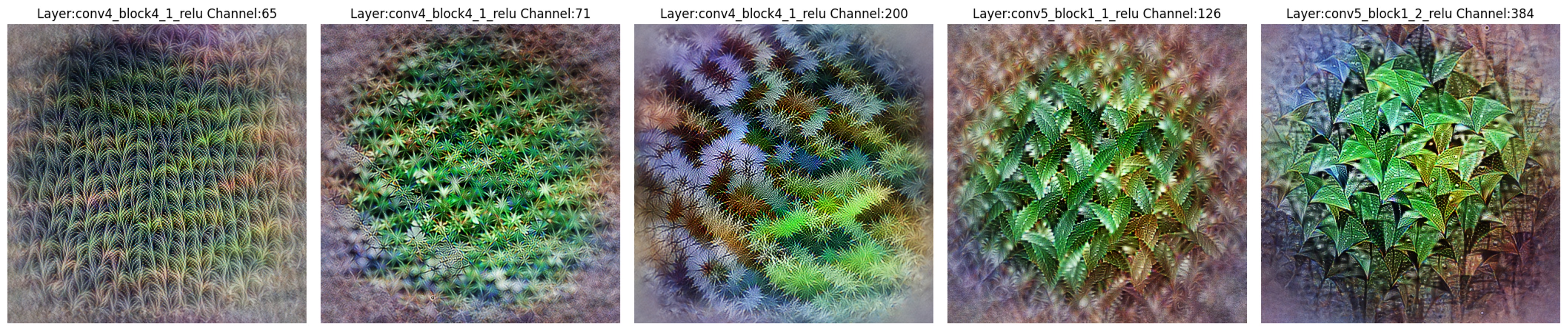}
    \includegraphics[width= 0.93\textwidth]{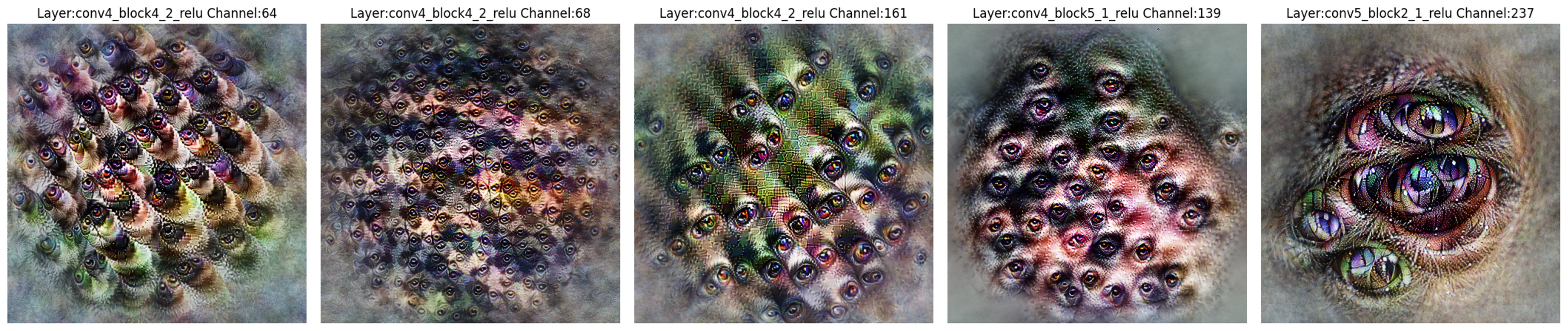}
    \includegraphics[width= 0.93\textwidth]{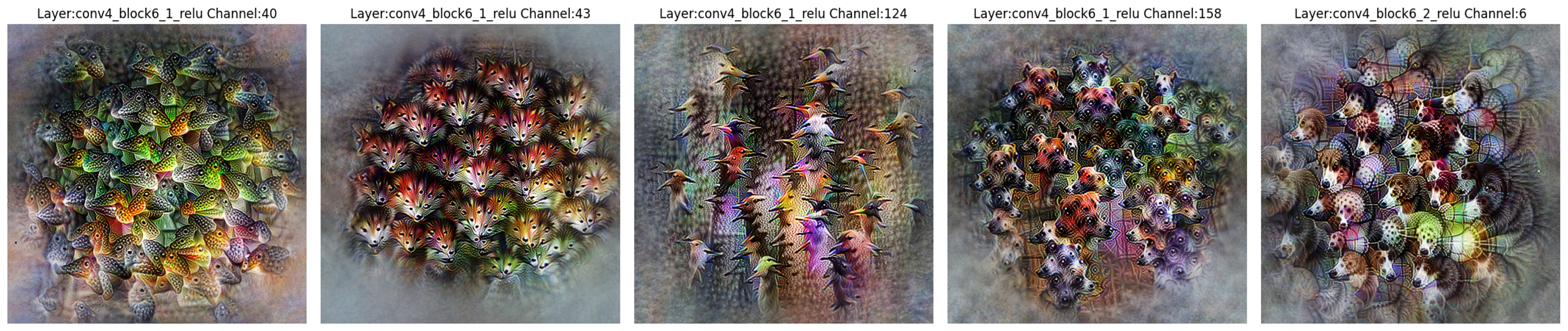}
    \includegraphics[width= 0.93\textwidth]{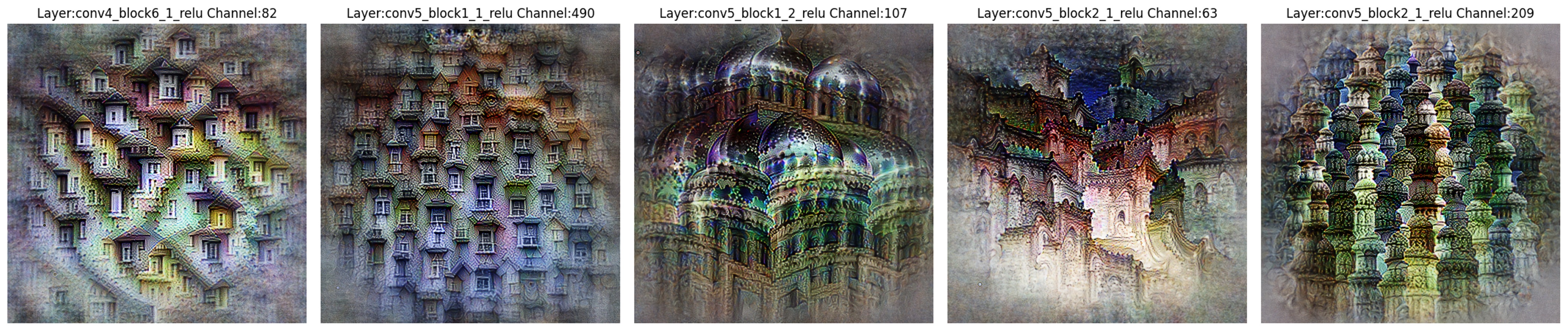}
    \caption{Visualization results of Resnet50V2 across different intermediate layers channels.}
    \label{fig:resnet50V2_ap_up}
\end{figure*}

\begin{figure*}
    \centering
    \includegraphics[width= 0.93\textwidth]{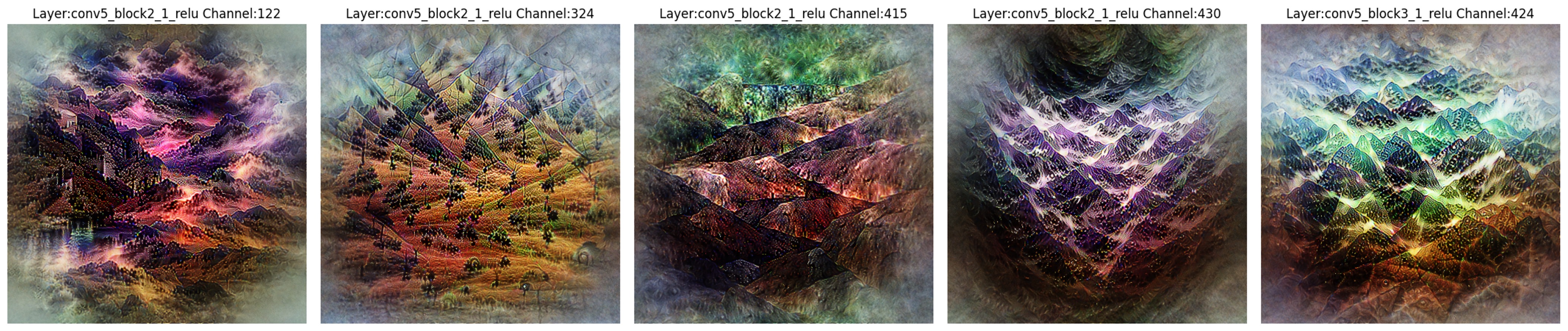}
    \includegraphics[width= 0.93\textwidth]{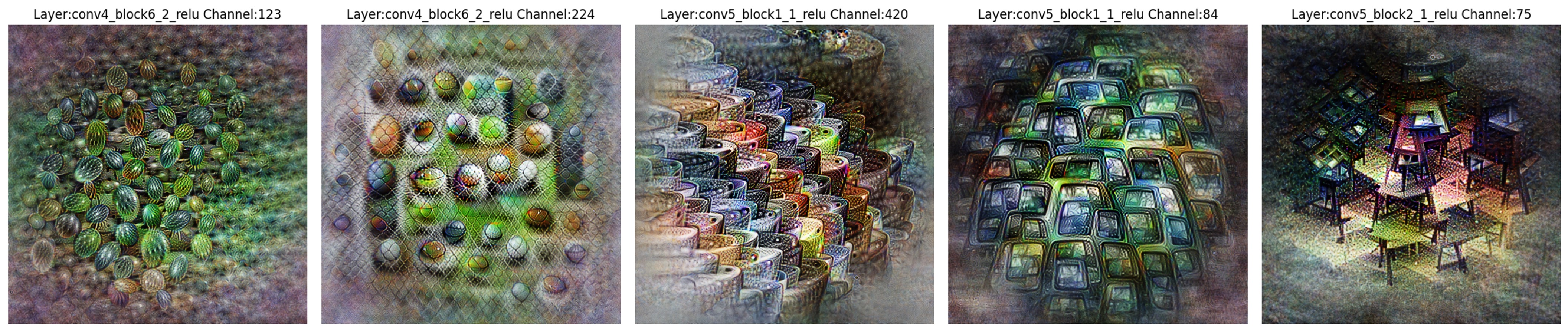}
    \includegraphics[width= 0.93\textwidth]{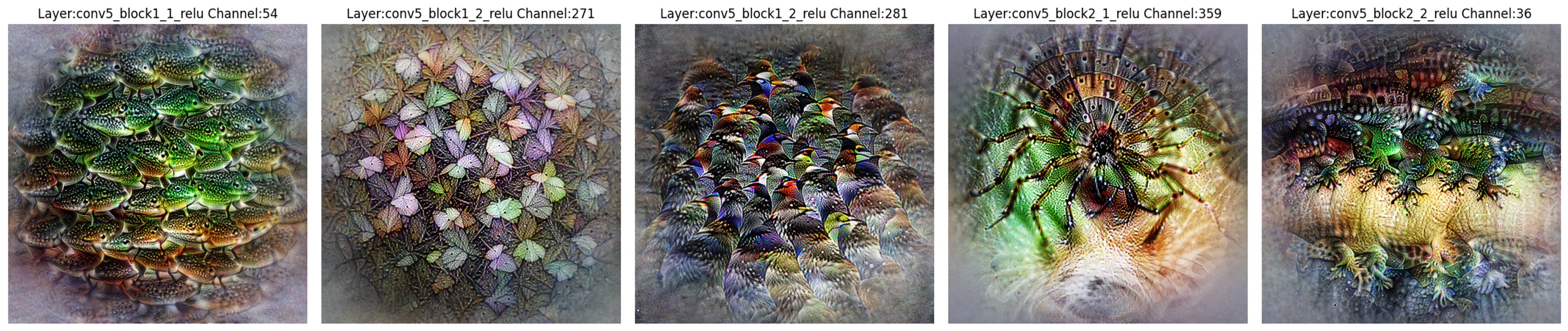}
    \caption{More visualization results of Resnet50V2 on deep layers channels.}
    \label{fig:resnet50V2_ap_down}
\end{figure*}


\end{document}